\documentclass[letterpaper, 10 pt, conference]{ieeeconf}  % Comment this line out if you need a4paper

\IEEEoverridecommandlockouts                              % This command is only needed if 
\usepackage{graphics} % for pdf, bitmapped graphics files
\usepackage{epsfig} % for postscript graphics files
\usepackage{amsmath} % assumes amsmath package installed
\usepackage{amssymb}  % assumes amsmath package installed
\usepackage{threeparttable}

\usepackage{graphicx}

\usepackage{verbatim}
\usepackage{epstopdf}
\usepackage{tabularx}
\usepackage{cite}
\usepackage{booktabs}
\usepackage{multirow}
\usepackage{color}
\usepackage{bm}

\title{\LARGE \bf
Learning Stereopsis from Geometric Synthesis for \\
6D Object Pose Estimation
}

\author{Jun Wu$^{1}$, Lilu Liu$^{1}$, Yue Wang$^{1}$ and Rong Xiong$^{1}$% <-this % stops a space
\thanks{$^{1}$Jun Wu, Lilu Liu, Yue Wang and Rong Xiong are with the State Key Laboratory
of Industrial Control Technology and Institue of Cyber-Systems and Control,
Zhejiang University, Zhejiang, China.}
\thanks{Corresponding author,
        {\tt\small wangyue@iipc.zju.edu.cn},
        Co-corresponding author,
        {\tt\small rxiong@zju.edu.cn}}.
}

\begin{document}

\maketitle
\thispagestyle{empty}
\pagestyle{empty}

%%%%%%%%%%%%%%%%%%%%%%%%%%%%%%%%%%%%%%%%%%%%%%%%%%%%%%%%%%%%%%%%%%%%%%%%%%%%%%%%
\begin{abstract}

Current monocular-based 6D object pose estimation methods generally achieve
less competitive results than RGBD-based methods, mostly due to the lack 
of 3D information. To make up this gap, this paper proposes a 3D geometric 
volume based pose estimation method with a short baseline two-view setting.
By constructing a geometric volume in the 3D space, 
we combine the features from two adjacent images to the same 3D space. 
Then a network is trained to learn the distribution of the position of 
object keypoints in the volume, and a robust soft RANSAC solver is deployed
to solve the pose in closed form. To balance accuracy and cost, 
we propose a coarse-to-fine 
framework to improve the performance in an iterative way. 
The experiments show that our method outperforms state-of-the-art 
monocular-based methods, and is robust in different objects and 
scenes, especially in serious occlusion situations.

\end{abstract}

%%%%%%%%%%%%%%%%%%%%%%%%%%%%%%%%%%%%%%%%%%%%%%%%%%%%%%%%%%%%%%%%%%%%%%%%%%%%%%%%
\section{Introduction}

6D object pose estimation aspires to estimate the rotation and translation of 
interested objects with regard to certain canonical coordinates. Accurate object 
pose estimation is the key to many real-world applications, such as robotic manipulation,
augmented reality, and human-robot interactions. This is a challenging problem due to the 
variety of objects appearance, occlusions between objects, and clutter in the scene. 

Based on the sensors they adopt, current object pose estimation methods can be
roughly categorized into two classes: 
monocular-based methods \cite{xiang2017posecnn}\cite{park2019pix2pose}\cite{li2018deepim}
and RGBD-based methods \cite{wang2019densefusion}\cite{wada2020morefusion}\cite{hua2021rede}. 
Previous researches have shown that the performance of monocular-based methods  
are generally less competitive than the other one. Without depth information, monocular-based methods
rely greatly on 2D image features to estimate 6D pose, thus involve more uncertainties.
Though monocular image lacks 3D awareness in nature, two or more frames of such images
combined together implicitly bring out the depth information, as has been verified in many
stereo or multi-view stereo tasks 
\cite{chang2018pyramid}\cite{li2019stereo}\cite{yao2018mvsnet}.  
However, when we are not certain about the object pose in one frame, unthinkingly linking 
another new frame with large baselines might introduce unknowable risks.
Thus, we opt to focus on estimating 6D object pose with a \emph{short-baseline two-view} setting, to 
lower the possibility of bringing more uncertainties. 

Intuitively, we consider two pipelines to tackle this problem: \emph{dense-depth} and \emph{sparse-depth}. 
\emph{Dense-depth} method aims to recover the dense depth map from two images, then solves the problem following 
current successful RGBD-based pose estimation methods. 
But under a short baseline setting, the parallax is probably too small to infer 
precise depth for all the points. On the other hand, \emph{sparse-depth} method only intends to obtain the depth of 
sparse keypoints useful for estimating object pose. For example, we can apply monocular-based methods to get
2D keypoints separately from the two images, then triangulate them to 3D space so as to establish 
the 3D-3D correspondence between scene and model. Nonetheless, this pipeline restores 3D points from two certain 
2D points, so is easy to be affected by the uncertainties from either estimation. 

Considering these two pipelines, we argue that under a short baseline,
restoring only several keypoints is more robust and efficient than recovering the 
whole 3D scene. But if we restore the sparse keypoint's depth 
by simply triangulating the already decided 2D keypoints, many precious 
early learned information is discarded, including semantic information
and context information. 
Therefore, instead of this \emph{sparse-depth late-fusion} 
pipeline, we attempt to keep as much information as possible to 3D, 
and directly learn 3D keypoints in geometric space.
Therefore, in this paper, we propose a 
\emph{sparse-depth early-fusion} mechanism to directly learn 3D 
keypoints by putting the 2D features to their corresponding 3D 
geometric position, trying to 
enhance the reliability and accuracy of RGB-based object pose estimation methods. 

In summary, the major contributions of this paper are as follows:

\begin{itemize}

\item We propose a novel \emph{sparse-depth early-fusion} framework 
for 6D object pose estimation in a \emph{short-baseline two-view} setting, 
which constructs a 3D geometric volume to learn the distribution of 
3D keypoints, and applies robust solver to estimate the pose.
\item We introduce a coarse-to-fine mechanism to predict 3D keypoints 
from the 3D geometric volume
by reducing the divergence between feature field and local keypoint heatmaps.
\item We show that our method is robust in different objects and 
scenes, especially in occlusion scenes, and effectively enhance the 
estimation accuracy
compared to monocular-based methods and our baselines.  

\end{itemize}

\begin{figure*}
        % \begin{center}
        \centering
        \includegraphics[width=0.9\textwidth]{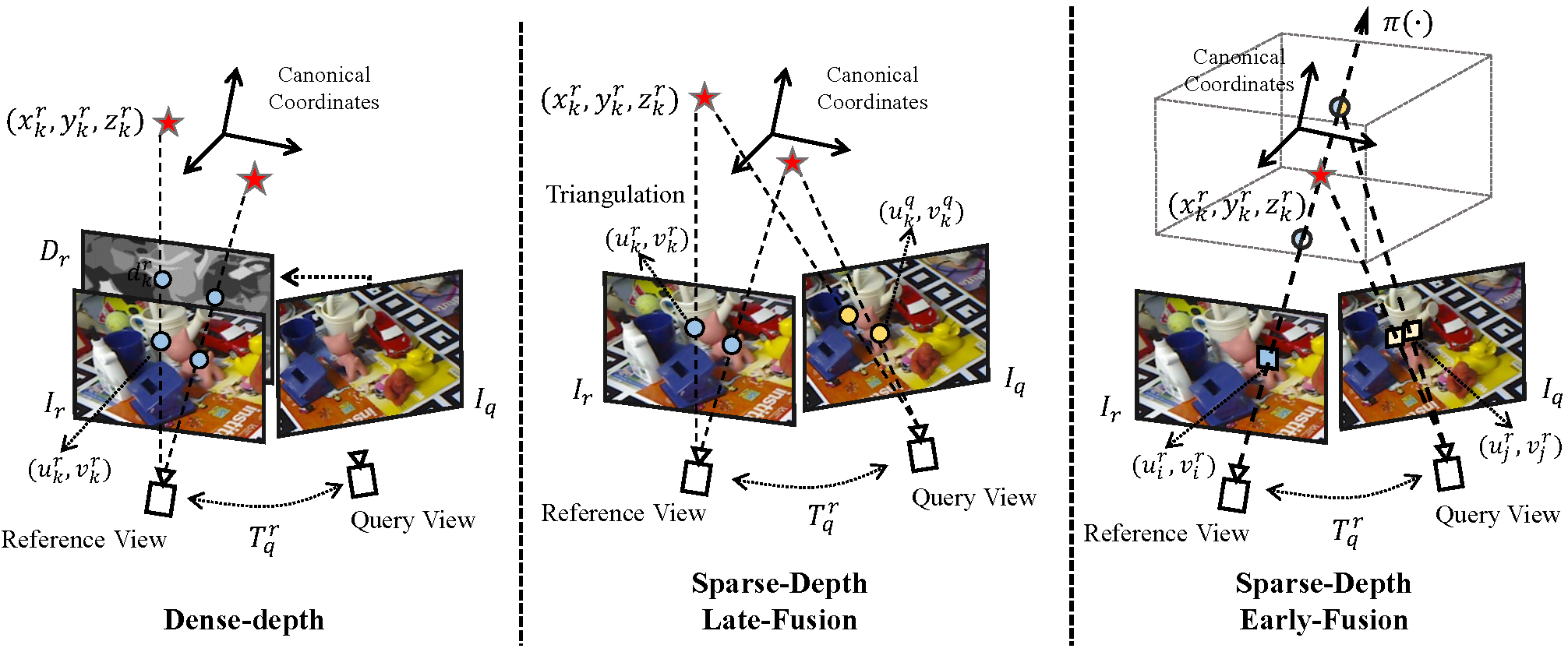}%\hfill
        \caption{Illustration of comparing 3 different pipelines to tackle two-view object pose 
                estimation problem. 
                \textbf{Left}: Dense-depth. 2D keypoints are computed in the reference view, and 
                a depth map is predicted by reference view and query view images. Together
                3D keypoints are predicted.
                \textbf{Middle}: Sparse-depth late-fusion. 2D keypoints are separately predicted 
                in both reference view and query view, then triangulated to 3D space to get
                3D keypoints.
                \textbf{Right}: Sparse-depth early-fusion. The images from reference view and 
                query view are combined together to directly predict 3D keypoints.}
        % \caption{(a) dense-depth (b) late-fusion sparse-depth (c) early-fusion sparse-depth(ours)}
        \label{fig:idea}
        % \end{center}
\end{figure*}

\section{Related Works}

\subsection{Single-view object pose estimation}

Most of the state-of-the-art approaches use single view observation to estimate pose.
Some methods take monocular images as input, and tackle this problem by establishing
2D-3D correspondence, followed by PnP algorithm to solve the pose 
\cite{park2019pix2pose}\cite{rad2017bb8}\cite{zakharov2019dpod}.  
Inspired by 2D object detection methods, \cite{tekin2018real} \cite{luo20203d} employs CNN 
to predict the 3D bounding box corners of the object in the image, and associate the 
corners with those in 3D CAD models to solve the pose. Since the corners are artificial, 
the estimation results are not satisfactory. 
To use more reliable correspondence, PVNet \cite{peng2019pvnet} selects keypoints from the object's model,
and train a CNN to predict the vertex from every pixel to those keypoints. 
Besides keypoints, HybridPose \cite{song2020hybridpose} also employs edge vectors and symmetry
correspondence to enrich the feature space for better estimation.

To further improve the estimation accuracy, another pipeline, RGBD-based methods
additionally deploy depth information 
\cite{wang2019densefusion}\cite{wada2020morefusion}\cite{he2020pvn3d}. 
Early research \cite{hinterstoisser2012model} 
\cite{hinterstoisser2011multimodal} compose contour vectors
from RGB image and surface normal vectors from depth image, and estimate the pose by 
template matching. Recently, PVN3D \cite{he2020pvn3d} uses a neural network to separately extract 
image and point features, then encode them pointwisely to vote for 3D keypoints, and solve 
the pose with 3D-3D correspondence. REDE \cite{hua2021rede} also encode multimodal features, and 
applies an outlier elimination mechanism to train the network in an end-to-end manner.
Because of the extra depth information, RGBD-based methods generally achieve better 
results than monocular-based methods. 

\subsection{Multi-view object pose estimation}

The performance of single-view pose estimation methods are relevant to the quality of 
the query observation to a large extend. Occlusions, poor lighting conditions, and 
textureless object surfaces are all possible factors to affect the results.
Hence, some researches involve more views of observations to 
estimate the pose\cite{collet2010efficient}\cite{collet2011moped}.
KeyPose \cite{liu2020keypose} uses stereo images to generate depth maps and uplifts detected 
2D keypoints to 3D space, so as to estimate the pose of textureless transparent objects.
\cite{li2018unified} first estimates the pose in every single view as hypothesis, 
then votes for the hypothesis by measuring their relative discrepancy. 
Moreover, CosyPose \cite{labbe2020cosypose} establish the consistency of objects across different
views, then refine the pose with consistency and relative camera pose together. 
However, CosyPose still takes the pose estimated from every single view, and applies refinement 
in a backend manner. Our method differs from it in estimating the pose from both the
query view and reference view in a frontend manner.

\begin{figure*}
        \begin{center}
        % \centering
        \includegraphics[width=0.96\textwidth]{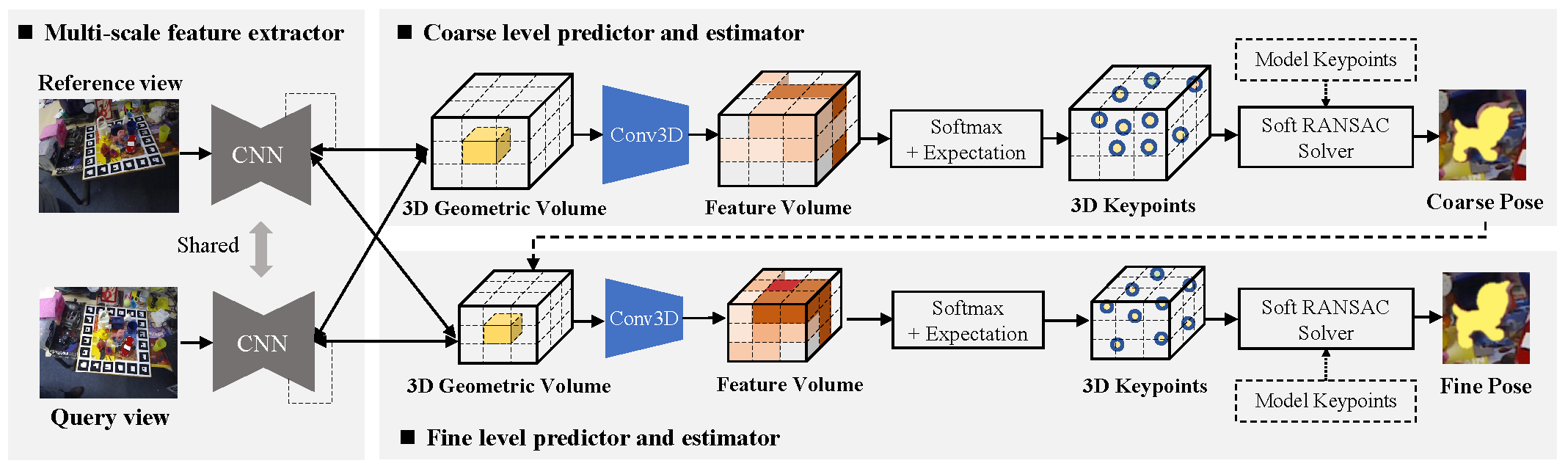}%\hfill
        \caption{\textbf{Method Overview.} Given two RGB images from reference view
        and query view, two feature maps are extracted by a 
        multi-scale convolutional neural network. Then a 3D geometric volume is 
        built surrounding an initial pose guess, which is then served to a 3D 
        convolutional network to learn the feature distribution volume. The feature 
        volume is then regularized to get the expected maximal value to predict 
        3D keypoints. Last, a soft RANSAC solver is deployed to solve the pose.
        The whole framework is conducted in a coarse-to-fine manner. 
        The components covered in blue need to be trained.}
        % \caption{(a) dense-depth (b) late-fusion sparse-depth (c) early-fusion sparse-depth(ours)}
        \label{fig:framework}
        \end{center}
\end{figure*}

\section{Method}

In this section, we introduce our overall pipeline, as illustrated in Fig.~\ref{fig:framework}. 
Taking the input reference image and query image, we extract 2D image features with a multi-scale
feature extractor network. Then, we build a 3D geometric volume to learn the feature distribution
in the 3D world space. By regularizing the distribution, we localize the 3D keypoints. Last, we 
adopt a soft RANSAC solver to solve the pose by 3D-3D correspondence. The whole process is iterated
in a coarse-to-fine framework. 

\subsection{Multi-layer Feature Extraction} 

Given two input images of size $W \times H$, we use $I_{r}$ to denote the reference image and $I_{q}$ to
denote the query image. 
Before applying keypoint prediction and pose estimation, we need to extract pixel-wise 
features from the input images. Since feature extraction is not the focus of this paper, we 
follow \cite{peng2019pvnet} to build a multi-scale convolutional neural network, to extract features
at multiple resolutions. 
Because precise keypoint prediction requires both high-level semantic features and 
low-level context features, we adopt the output of the last upsampling layer of size 
$W \times H$, together with the output of the third from the last upsampling layer of size
$W/4 \times H/4$, to build our 3D geometric volume. Also, to force the network to 
focus on the object rather than the whole scene, we take the output of the segmentation 
branch of the extractor as our predicted mask, and use these two masks to build 
the volume, too.

\subsection{Constructing 3D Geometric Volume} 

To learn the 3D keypoint distribution, we directly build a 3D geometric volume
in the 3D space. Without loss of generality, we divided the interested 3D space 
into regular 3D grids of size $(H_{v},W_{v},D_{v})$, and the size of each grid is 
$(h_{v},w_{v},d_{v})$. The axes of the grid coordinates are centralized in an initial guess 
and parallel to the reference camera coordinates.

With known camera intrinsic parameters ${\{f_{x},f_{y},c_{x},c_{y}\}}$, 
we create a many to one projection from 3D grid 
space to image space by camera projection function $\pi(\cdot)$
\begin{equation}
z
\left(
\begin{array}{ccc}
u \\ v \\ 1
\end{array}
\right)
=
\left(
\begin{array}{ccc}
f_{x} & 0 & c_{x} \\
0 & f_{y} & c_{y} \\
0 & 0 & 1
\end{array}
\right)
\left(
\begin{array}{ccc}
x \\ y \\ z
\end{array}
\right)
\end{equation}
where ${(x,y,z)}$ is point in 3D grid space, and ${(u,v)}$ refers to image pixel position.
The grid space is also projected to the query image space by the relative
camera pose between the two views.
This projection is fully differentiable, so it's feasible to be included as part 
of our network. It can be implemented by bilinear interpolation.

Through this process, each grid in 3D space is related to a point
in the image space. Then we assign the high-level and low-level 
features extracted before of the 2D point to its related 3D point.
In this way, we uplift the 2D features to the 3D space without 
approximation. Points projected outside the image are assigned to initial feature values.

What's more, considering that not only the features themself, but also 
the relationship between two feature vectors of the same 3D point,  
comprise significant details of the object, we simply concatenate these two feature vectors 
, to keep as much information as possible for the following network to learn.

\subsection{Coarse-to-fine Keypoint Prediction}

Instead of regressing the geometric volume to get the 3D keypoints,
we propose to learn the distribution of keypoints by reducing the 
divergence between the feature field 
and the local keypoint heatmaps. The local keypoint heatmaps are defined as 
\begin{equation}
Q_{i}^{k}(p|m_{i},\theta) \sim \mathcal{N}(\theta m_{i},\sigma_{i})
\end{equation}
where $\{m_{i}\}_{i=1}^N$ refers to the model keypoints, $\theta$ is the 
target pose, and $\sigma_{i}$ is the hyperparameters. This distribution 
represents the local heatmap we expect to be highlighted as possible keypoints 
locations.

And the feature field is defined as
\begin{equation}
Q_{i}^{v}(p|V(\cdot)) = \sum_{j\in \Omega} w_{j}V([p_{j}])
\end{equation}
where $\{w_{j}\}_{j\in \Omega}$ are the trilinear interpolation coefficients, 
$[p_{j}]$ are the 8 neighbour grids of the conditioned position $p$, 
and $V(\cdot)$ is the value operation in the feature field. This distribution 
describes how likely a position in the field is to be the keypoints.

To minimal the divergence of these two distribution, we adopt a Kullback-Leibler
divergence loss as 
\begin{equation}
Loss_{KL} = D_{KL}(Q^{v}\|Q^{k}) = -\sum_{i=1}^{N} Q_{i}^{v}log(\frac{Q_{i}^{k}}{Q_{i}^{v}})
\end{equation} 

To help the feature volume to learn the divergence, we build a simple 3-layers 3D convolutional 
neural network to intensify the difference, and add a softmax layer 
to regularize the output volume to further concentrate the distribution.
After this process, the feature field is supposed to embody the distribution possibility of the expected keypoints. Hence, 
we maximize the marginal possibility of each dimension to find the optimal
keypoint locations, and employ a smooth ${L_{1}}$ loss to evaluate the keypoints
\begin{equation}
Loss_{kpt} = \frac{1}{N}\sum_{i}^{N}smooth_{L_{1}}(p_{i},\hat{p_{i}})
\end{equation}

It's reasonable to 
assume that finer grid resolutions bring better prediction results.
But higher resolution also costs more computation resources.
Therefore, we employ a coarse-to-fine strategy. At the first coarse level, 
we build the 3D geometric volume with a resolution of $1cm$ and a larger 
space range. While at the second finer level, we decrease the size of 
each grid to $0.5cm$ and reduce the space range.

\begin{figure}
        % \begin{center}
        \centering
        \includegraphics[width=0.3\textwidth]{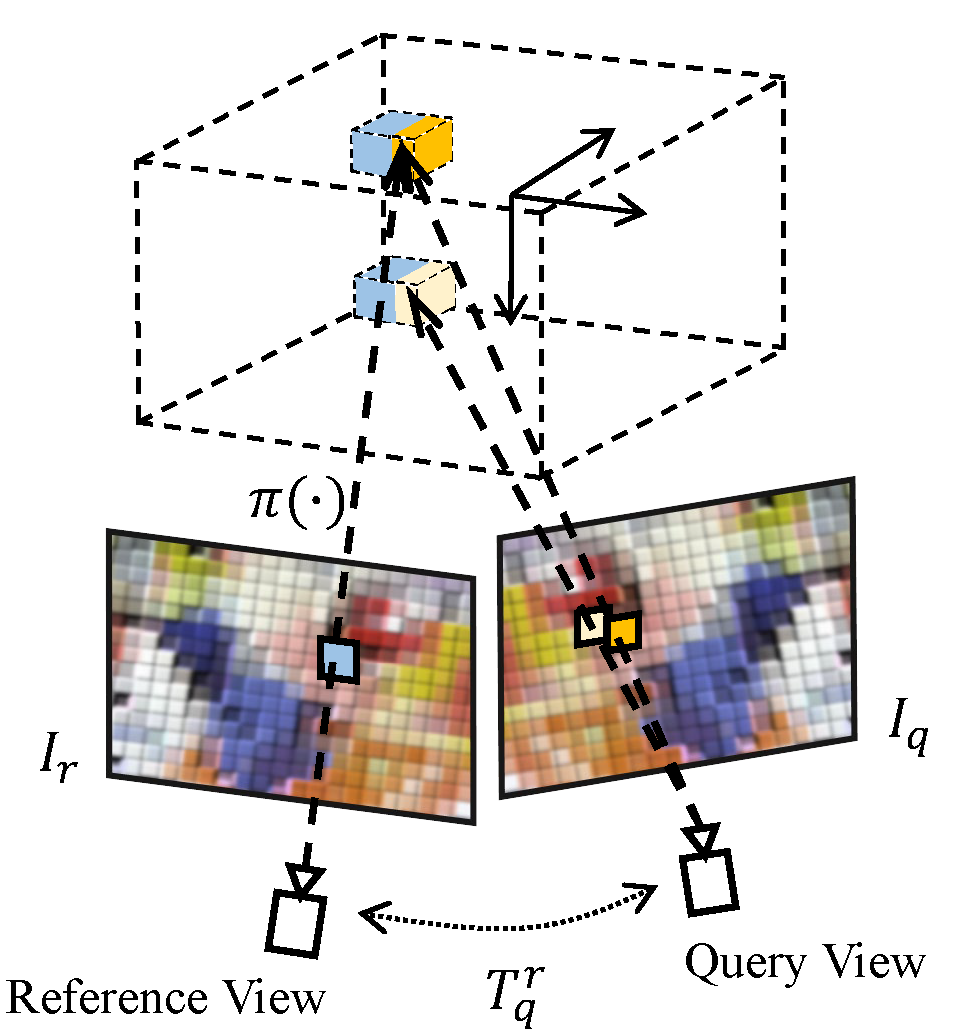}%\hfill
        \caption{\textbf{3D Geometric Volume.} A 3D geometric volume is constructed upon 
        two feature maps. The 3D space is divided into regular grids. 
        By camera projection matrix, each grid 
        is projected back to 2D image space to get its related pixel. 
        The two feature vectors of the two related pixels are concatenated 
        together as the 3D grid's feature.}
        % \caption{(a) dense-depth (b) late-fusion sparse-depth (c) early-fusion sparse-depth(ours)}
        \label{fig:3DGV}
        % \end{center}
\end{figure}

\subsection{Soft Robust Solver}

Given the predicted 3D keypoints $\{\hat{p}_{i}\}_{i=1}^N$ and the model keypoints 
$\{m_{i}\}_{i=1}^N$, we then solve the object pose by minimizing the distance 
between predicted points and model points. The optimization problem is 
\begin{equation}
\hat{\theta} = \arg\min_{\theta}\|\theta m_{i}-\hat{p}_{i}\|_{2}
\end{equation}        
where $\theta$ represents the 6D object pose.

Though this optimization problem can be solved by SVD in closed form, it is 
easy to be disturbed by outliers. Many recent works adopt RANSAC(RANdom SAmple Consensus)
algorithm \cite{fischler1981random} or its variants to pick inliers 
\cite{peng2019pvnet}\cite{he2020pvn3d}. 
However, in our case, we only 
have $N$ 3D points for optimization, the extra information needed to perform RANSAC 
is absent. Thus, directly applying the classic RANSAC algorithm is barely useful. 
Another reasonable approach is to apply a nonlinear optimization solver 
to update the pose in the feature field, such as Gauss-Newton algorithm 
\cite{hartley1961modified} or Levenberg-Marquardt algorithm\cite{more1978levenberg}.
But we find that the results are not as good as expected, probably
because the distribution of the feature field is not convex everywhere, 
which leads to an incorrect updating orientation or excessive step.
Therefore, we propose a soft RANSAC solver to softly count the inliers to solve a more robust pose.

Taken the predicted keypoints $\{\hat{p}_{i}\}_{i=1}^N$, we calculate all possible 
poses with every 3 points by SVD, which brings us a set of pose hypothesis
${\{\theta_{k}\}_{k=1}^{{C_{N}^{3}}}}$. For each hypothesis, 
we evaluate the distance for every predicted and model keypoints under the hypothesis pose
\begin{equation}
d_{k,i} = \|\theta_{k}m_{i}-\hat{p}_{i}\|_{2}
\end{equation}       
Then, instead of masking out the inliers with a hard threshold, we employ a sigmoid operation 
to softly classify the keypoints, and sum up the outputs of sigmoid as the soft inlier count 
of this hypothesis
\begin{equation}
Score_{k} = \sum_{i}^{N}sigmoid(\gamma_{1}(-d_{k,i}+\gamma_{2}))
\end{equation}
Last, we regularize the scores for all hypothesis by softmax, and softly aggregate them into
a final pose 
\begin{equation}
\hat{\theta} = \sum_{k}^{K}Score_{k} \cdot \theta_{k}
\end{equation}

The loss to evaluate the predicted pose is 
\begin{equation}
Loss_{pose} = \|\hat{t}-t\|_{2}+\alpha \|\hat{R}R^{T}-I\|_{F}
\end{equation}

In total, our network is trained with joint loss from both coarse level and fine level
\begin{equation}
Loss = \sum_{j}\beta_{1}Loss_{pose_{j}}+\beta_{2}Loss_{kpt_{j}} +\beta_{3} Loss_{KL_{j}}
\end{equation}

\section{Experiments}

In this section, we evaluate the proposed method by comparing it with 
the state-of-the-art methods on the Occlusion LineMOD dataset\cite{brachmann2014learning},
to validate our strength in tackling occluded scenes using only RGB inputs.

\subsection{Dataset}

Occlusion LineMOD\cite{brachmann2014learning} dataset is a widely used benchmark 
for object 6D pose estimation task with serious occlusion. 
It contains 8 objects from the LineMOD dataset\cite{hinterstoisser2011multimodal},
but includes more challenges such as low resolution, cluttered scenes, and severe
occlusions. In a lot of scenes of Occlusion LineMOD, 
only a small part of the object can be observed. 
Therefore, it's a suitable dataset to evaluate our method.
All the data is used for testing. The model is trained on LineMOD dataset. 
During training and testing, we pair up every two adjacent frames in the test list. 
The average relative camera distance of all the testing pairs is $0.168m$, and the minimal distance 
is $0.004m$.

\subsection{Metrics}

In object pose estimation task, the most commonly used metrics are ADD \cite{hinterstoisser2012model}
and ADD-S \cite{xiang2017posecnn}.
Given an object model of $M$ points, 
ADD metric evaluates the average distance between model points transformed with 
predicted and ground truth pose respectively 
\begin{equation}
ADD = \frac{1}{M}\sum_{i=1}^{M}\|\theta p_{i}-\hat{\theta} p_{i}\|_{2}
\end{equation}

While ADD-S metric calculates the average distance between the closest points, which 
is used for evaluating symmetric object
\begin{equation}
ADD\mbox{-}S = \frac{1}{M}\sum_{i=1}^{M}\min_{j\in M}\|\theta p_{j}-\hat{\theta} p_{i}\|_{2}
\end{equation}

We use ADD for non-symmetric objects and ADD-S for symmetric objects. 
An estimation is regarded successful if the ADD(-S) is less than 
$10\%$ of the object's diameter. 

\subsection{Implementation Details}

In inplementation, we follow \cite{peng2019pvnet} to select 9 keypoints for 
every object, and perform the same data augmentation.

In the coarse level, we compute the center pixel position in the predicted 
mask in reference view, and unproject it to a 3D position with a prior depth,
which is the average of the object depths in the dataset.
Then, we build the 3D geometric volume around this initial position 
in range $[-0.3,0.3]\times [-0.3,0.3]\times [-0.3,0.3]$(meters), 
with grid size of $0.01m$.
The keypoint prediction network contains three 3D convolutional layers, each followed with
a 3D BatchNorm layer and a ReLU layer, one output 3D convolutional layer, and finally 
a LogSoftmax layer.

In the fine level, we take the estimated position from the coarse level as
the initial position, and build the 3D geometric volume around it with grid size 
of $0.005m$. The range of the fine volume is dependent on the diameter of each object.
The keypoint prediction network is nearly the same as the network used 
in the coarse level, but contains one less convolutional layer.

We run all our training and experiments on a machine equipped with an 
Intel(R) Xeon(R) Silver 4216 CPU at 2.10GHz, and an NVIDIA GeForce RTX 3090 GPU.

% \begin{table*}[htbp]
%         \centering
%         \caption{Performance comparison on Occlusion LineMOD (ADD(-S)$<0.1d$).}
%             \begin{threeparttable}
%             \begin{tabular}{lccccccc}
%                 \toprule
%                 & RGB & & & & 2-view RGB & RGBD & \\
%                 \midrule
%                 & PVNet & Pix2Pose & DPVL & HybridPose & Ours & REDE & PVN3D  \\
%                 \midrule
%                 \midrule
%                 ape   & 15.8 & 22.0 & 19.2 & 20.9 & \textbf{37.2} & 56.1 & 33.9  \\
%                 can   & 63.3 & 44.7 & 69.8 & \textbf{75.3} & 63.5 & 80.7 & 88.6   \\
%                 cat   & 16.7 & 22.7 & 21.1 & \textbf{24.9} & 21.8 & 23.3 & 39.1 \\
%                 driller & 65.7 & 44.7 & 71.6 & \textbf{70.2} & 67.2 & 85.0 & 78.4\\
%                 duck & 25.2 & 15.0 & 34.3 & 27.9 & \textbf{36.9} & 47.0 & 41.9  \\
%                 eggbox* & 50.2 & 25.2 & 47.3 & \textbf{52.4} & 42.3 & 43.7 & 80.9 \\
%                 glue* & 49.6 & 32.4 & 39.7 & 53.8 & \textbf{62.2} & 71.3 & 68.1 \\
%                 holepuncher & 39.7 & 49.5 & 45.3 & \textbf{54.2} & 43.7 & 62.6 & 74.7 \\
%                 % \midrule
%                 % average & 40.8 & 32.0 & 47.5 &  & 58.5 & 63.2 \\
%                 \bottomrule
%             \end{tabular}%
%             \begin{tablenotes}
%                 \item *denotes symmetric objects.
%             \end{tablenotes}
%             \end{threeparttable}
%         \label{tab:occ}%
% \end{table*}%

\begin{table}[htbp]
        \centering
        \caption{Performance comparison on Occlusion LineMOD (ADD(-S)$<0.1d$).}
            \begin{threeparttable}
            \begin{tabular}{lccccc}
                \toprule
                & RGB & & & 2-view RGB & \\
                \midrule
                & PoseCNN & PVNet & Hu & Late-fusion & Ours  \\
                & \cite{xiang2017posecnn} & \cite{peng2019pvnet} & \cite{hu2019segmentation} & & \\
                \midrule
                \midrule
                ape & 9.6 & 15.8 & 12.1 & 34.5  &  \textbf{37.2}  \\
                can & 45.2 & 63.3 & 39.9 & 57.9 & \textbf{64.6}   \\
                cat & 0.93 & 16.7 &  8.2 & \textbf{24.3} & 22.8 \\
                driller & 41.4 & 65.7 & 47.1  & 58.3 &  \textbf{67.2} \\
                duck & 19.6 & 25.2 & 11.0 & 33.5 &  \textbf{36.9}   \\
                eggbox* & 22.0 & \textbf{50.2} & 24.7 & 46.3 &  42.3  \\
                glue* & 38.5 & 49.6 & 39.5 & 60.0 & \textbf{62.2}  \\
                holepuncher & 22.1 & 39.7 &  21.9 & 41.6 & \textbf{43.7}  \\
                \midrule
                average & 24.9 & 40.8 & 27.0 & 44.6 & \textbf{47.1}  \\
                \bottomrule
            \end{tabular}%
            \begin{tablenotes}
                \item *denotes symmetric objects.
            \end{tablenotes}
            \end{threeparttable}
        \label{tab:occ}%
\end{table}%

\begin{table}[htbp]
        \centering
        \caption{Comparison of Average 3D Keypoint Prediction Error.}
            \begin{threeparttable}
            \begin{tabular}{lccc}
                \toprule
                & Late-fusion & REDE\cite{hua2021rede} & Ours \\
                % & & \cite{hua2021rede} & \\
                \midrule
                \midrule
                ape   & 0.121 & \textbf{0.030} & 0.043 \\
                can   & 0.048 & 0.034 & \textbf{0.032} \\
                cat   & 0.261 & \textbf{0.090} & 0.114 \\
                driller & 0.065 & 0.041 & \textbf{0.029} \\
                duck & 0.176 & \textbf{0.034} & 0.035 \\
                holepuncher & 0.084 & 0.058 & \textbf{0.027} \\
                \midrule 
                average & 0.126 & 0.048 & \textbf{0.047} \\
                \bottomrule
            \end{tabular}%
            \end{threeparttable}
        \label{tab:kpterror}%
\end{table}%

\subsection{Results on Benchmark Dataset}

We evaluate the performance of our method in Occlusion LineMOD dataset, to 
compare with state-of-the-art monocular-based methods and the other 2-view method 
late-fusion, as shown in Table.~\ref{tab:occ}. 
Since the late-fusion approach hasn't been explored by other works to our best knowledge, 
we implement the approach by ourselves. For fair comparison, 
we take the 2D predicted keypoints from PVNet \cite{peng2019pvnet},
and triangulate the two keypoints to 3D space by classic method \cite{bradski2000opencv}. 
By doing so, the late-fusion approach shares the same feature maps with our proposed method,
and the major difference lies in the keypoint prediction mechanism. 

Pose estimation under serious occlusion situations is a difficult task,
especially for monocular-based methods, thus
not a lot of current methods report their performance in this dataset. 
As the table shows, 2-view based methods generally acquire better results than the monocular-based 
methods, illustrating the benefits of introducing stereopsis in the estimation task.
Compared to the other 2-view method late-fusion, our method exceeds with a margin of 
$3.5\%$, which indicates the advantage of early geometric fusion of 2D features.
Overall, our method achives the best performance in 6 classes and gets the highest 
average recall. 
% Compared to \cite{pvnet} which shares exactly the same feature 
% maps with us, our method outperforms in all classes except for eggbox, 

\begin{figure}[htbp]
        % \begin{center}
        \centering
        \includegraphics[width=0.45\textwidth]{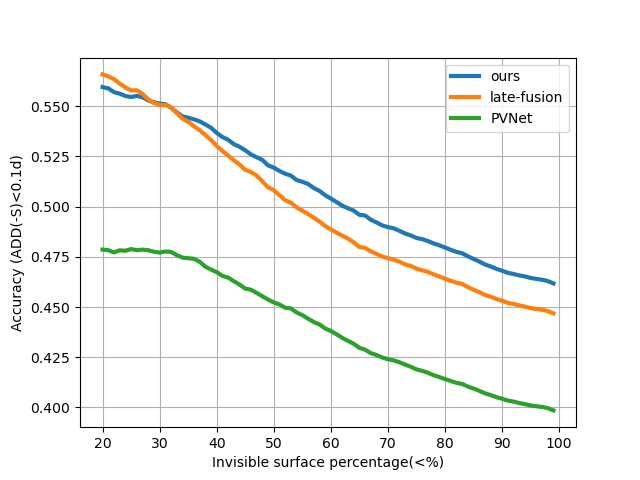}%\hfill
        \caption{Performance with respect to invisible surface percentage.
        Invisible surface percentage is computed as the ratio between 
        invisible mask area whole mask area. 
        The proposed method is more robust under heacy occlusion situation.}
        % \caption{(a) dense-depth (b) late-fusion sparse-depth (c) early-fusion sparse-depth(ours)}
        \label{fig:occlevel}
        % \end{center}
\end{figure}

To verify our robustness towards occlusion, we also draw
accuracy curve under increasing levels of invisible surfaces on
Occlusion LineMOD dataset. Following \cite{wang2019densefusion}, 
levels of occlusion are measured by calculating the invisible surface
percentage of model points projected in the image frame. 
We measure the performance 
in the whole Occlusion LineMOD dataset.
The accuracy of ADD(-S) smaller than $0.1d$ curve is shown in Fig.~\ref{fig:occlevel}, 
it can be seen that our performance under occlusion is more stable compared to 
monocular-based methods and the late-fusion approach,
especially in heavy occlusion situations.

What's more, we present the average 3D keypoint prediction L2 error in Table.~\ref{tab:kpterror}.
The error is calculated between our predicted scene 3D keypoints and ground truth 
scene keypoints with an L2 distance. We compare the distance with the late-fusion 
approach and a RGBD-based method \cite{hua2021rede}.
Except for symmetric objects, our method predicts 3D keypoints with a slightly better accuracy
than the RGBD-based methods \cite{hua2021rede}, and far better than the late-fusion approach with 
a margin of $0.079m$.

\begin{figure}
        \begin{center}
        % \centering
        \includegraphics[width=0.48\textwidth]{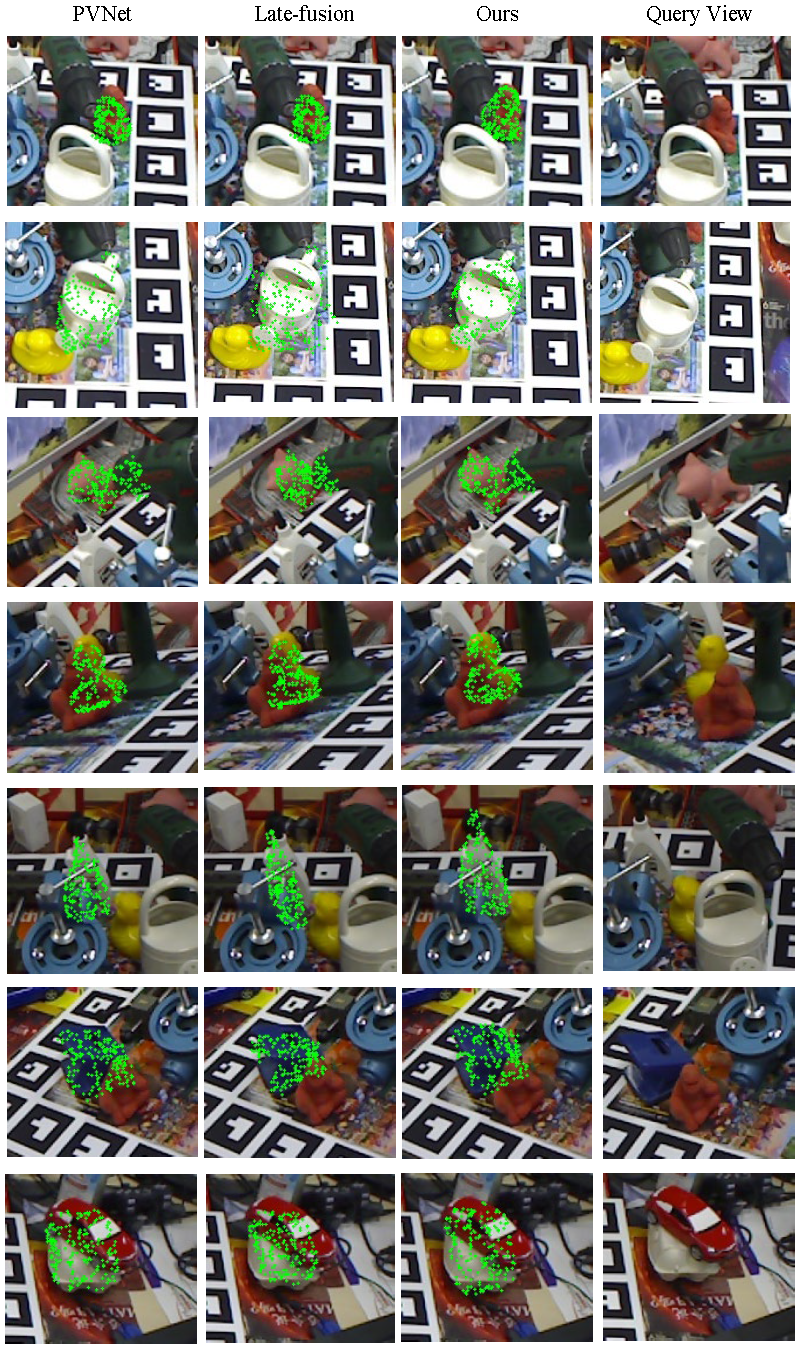}%\hfill
        \caption{\textbf{Visualization Results Samples.} Some visualization 
        results on OCC-LM dataset. All the points are projected to the reference 
        image by estimated pose of different methods. The last column 
        shows the query view used in our proposed method. All images are cropped
        for better visualization, and the query view is cropped with the same position
        and size as the reference view to show the parallax. The last row
        shows a failure case in eggbox class.}
        % \caption{(a) dense-depth (b) late-fusion sparse-depth (c) early-fusion sparse-depth(ours)}
        \label{fig:results}
        \end{center}
\end{figure}

Some visualization results are shown in Fig.~\ref{fig:results}. 
We project the object CAD model transformed by
the estimated pose and draw the points on the reference view.
All the images are cropped for better visualization, and the query view 
is cropped with the same position and size to show the parallax.
It can be observed that compared with \cite{peng2019pvnet} and late-fusion approach,
our method can accurately estimate the pose of objects,
especially in some hard cases with occlusion. 

\begin{table}[htbp]
        \centering
        \caption{Ablation Study on Coarse-to-Fine Mechanism.}
            \begin{threeparttable}
            \begin{tabular}{lcccc}
                \toprule
                & coarse & & coarse+fine & \\
                \midrule
                & ADD(-S) & $<$0.1d & ADD(-S) & $<$0.1d \\
                & (avg/med) & & (avg/med) & \\
                \midrule
                \midrule
                ape   & 0.045/0.017 & 29.3 & $\downarrow0.005/0.003$ & $\uparrow7.9$ \\
                can   & 0.032/0.017 & 55.9 & $\downarrow0.005/0.003$ & $\uparrow8.7$ \\
                cat   & 0.128/0.096 & 15.3 & $\downarrow0.026/0.039$ & $\uparrow7.5$\\
                driller & 0.032/0.020 & 61.4 & $\downarrow0.004/0.002$ & $\uparrow5.8$ \\
                duck & 0.040/0.019 & 28.9 & $\downarrow0.005/0.003$ & $\uparrow8.0$\\
                eggbox* & 0.096/0.037 & 32.7 & $\downarrow0.015/0.015$ & $\uparrow9.6$ \\
                glue* & 0.066/0.014 & 55.9 & $\downarrow0.004/0.003$ & $\uparrow6.3$ \\
                holepuncher & 0.030/0.018 & 39.4 & $\downarrow0.003/0.002$ & $\uparrow4.3$ \\
                \midrule 
                average & & 39.9 & & $\uparrow7.2$ \\
                \bottomrule
            \end{tabular}%
            \begin{tablenotes}
                \item *denotes symmetric objects.
            \end{tablenotes}
            \end{threeparttable}
        \label{tab:ablation}%
\end{table}%

\subsection{Ablation Study}

To validate the proposed coarse-to-fine mechanism, we compare the results after only 
coarse level network to the final coarse-to-fine network. 
The experiments are conducted in Occlusion LineMOD dataset.
Table.~\ref{tab:ablation} summarize the results. 
With the help of the coarse-to-fine mechanism, 
the recall of every class is increased with an average improvement of $7.2\%$.
Also, the average and medium ADD are all decreased, especially in hard classes such 
as cat and eggbox, with reductions of $0.026/0.039$ and $0.015/0.015$ respectively. 
The results verify the benefits of learning the keypoint 
distribution from coarse resolution to finer resolution, 
and the ability of the mechanism to improve an incorrect initial guess to a certain 
extent.

% \textbf{performance in different camera relative baselines}
% To better illustrate the performance of our method under different camera view distance, 
% we show the ADD(-S) with respect to the distance between the two camera centers in Fig

\section{CONCLUSIONS}
In this paper, we propose an geometric volume fusion based object 6D pose estimation method under 
short-baseline two-view setting. We build a geometric volume in 3D space to 
restore the object's 3D information from two monocular RGB images. By regularization and 
learning, the volume is trained to highlight the position of object keypoints. A soft ransac 
solver is deployed to solve the pose in closed form.
In addition, we deploy a coarse-to-fine framework to improve the estimation accuracy. 
Experiments show that our method outperforms the state-of-the-art monocular-based 
methods and our baselines in serious occlusion datasets.


%%%%%%%%%%%%%%%%%%%%%%%%%%%%%%%%%%%%%%%%%%%%%%%%%%%%%%%%%%%%%%%%%%%%%%%%%%%%%%%%

%%%%%%%%%%%%%%%%%%%%%%%%%%%%%%%%%%%%%%%%%%%%%%%%%%%%%%%%%%%%%%%%%%%%%%%%%%%%%%%%

%%%%%%%%%%%%%%%%%%%%%%%%%%%%%%%%%%%%%%%%%%%%%%%%%%%%%%%%%%%%%%%%%%%%%%%%%%%%%%%%

%%%%%%%%%%%%%%%%%%%%%%%%%%%%%%%%%%%%%%%%%%%%%%%%%%%%%%%%%%%%%%%%%%%%%%%%%%%%%%%%
% \clearpage

\bibliographystyle{ieeetr}
\bibliography{reference.bib}

\end{document}